\documentclass{article}

\usepackage{xcolor}
\usepackage{graphicx}
\usepackage{times}
\usepackage{latexsym}
\usepackage{microtype}
\usepackage{amsfonts}
\usepackage{amsthm}
\usepackage{amsmath}
\usepackage{amssymb}
\usepackage{mathtools}
\usepackage{graphicx}
\usepackage{subfigure}
\usepackage{booktabs}
\usepackage{multirow}
\usepackage{url}
\usepackage{hyperref}
\usepackage{csquotes}
\usepackage{enumitem}
\usepackage{subcaption}

\usepackage[accepted]{icml2024} 

\DeclareMathOperator{\Law}{Law}

\theoremstyle{plain}

\theoremstyle{definition}

\theoremstyle{remark}

\icmltitlerunning{Improving Diffusion Models's Data-Corruption Resistance using Scheduled Pseudo-Huber Loss}

\begin{document}

\twocolumn[

\icmltitle{Improving Diffusion Models's Data-Corruption Resistance using Scheduled Pseudo-Huber Loss}

\icmlsetsymbol{equal}{$^{\bigotimes\nolimits}$}

\begin{icmlauthorlist}
\icmlauthor{Artem Khrapov}{hw}
\icmlauthor{Vadim Popov}{hw}
\icmlauthor{Tasnima Sadekova}{hw}
\icmlauthor{Assel Yermekova}{hw}
\icmlauthor{Mikhail Kudinov}{hw}
\end{icmlauthorlist}

\icmlaffiliation{hw}{Huawei Noah's Ark Lab, Moscow, Russia}

\icmlcorrespondingauthor{Artem Khrapov}{artemkhrapov2001@yandex.ru}

\icmlkeywords{diffusion probabilistic modeling, text-to-image, text-to-speech, data corruption, robustness, Huber loss}

\vskip 0.3in
]

\printAffiliationsAndNotice{}

\begin{abstract}

Diffusion models are known to be vulnerable to outliers in training data. In this paper we study an alternative diffusion loss function, which can preserve the high quality of generated data like the original squared $L_{2}$ loss while at the same time being robust to outliers. We propose to use pseudo-Huber loss function with a time-dependent parameter to allow for the trade-off between robustness on the most vulnerable early reverse-diffusion steps and fine details restoration on the final steps. We show that pseudo-Huber loss with the time-dependent parameter exhibits better performance on corrupted datasets in both image and audio domains. In addition, the loss function we propose can potentially help diffusion models to resist dataset corruption while not requiring data filtering or purification compared to conventional training algorithms.

\end{abstract}

\section{Introduction}
\label{sec:intro}

Over the past few years denoising diffusion probabilistic models \cite{DDPM, Song-main} have been achieving remarkable results in various generative tasks. It is no exaggeration to say that almost all common computer vision generative problems such as conditional and unconditional image generation \cite{DPMs-beat-GANs}, image super-resolution \cite{image-superres}, deblurring \cite{DDRM}, image editing \cite{sdedit} and many others found high-quality solutions relying on diffusion models. Such success inspired researches specializing in other fields to apply diffusion-based approaches to the tasks they worked on. As a result, there appeared diffusion models solving a vast range of tasks, e.g. text-to-speech synthesis \cite{Grad-TTS}, music generation \cite{music-generation}, audio upsampling \cite{nuwave2}, video generation \cite{videofusion}, chirographic data generation \cite{chirodiff} and human motion generation \cite{human-motion} to name but a few. In several areas like text-to-image \cite{stable-diffusion} and text-to-video \cite{videofusion} generation diffusion models lead to breakthroughs and became \textit{de-facto} a standard choice.

Various aspects related to diffusion models attracted interest of specialists in generative modeling. Soon after diffusion models had been introduced, numerous attempts were made to accelerate them --- either by investigating differential equation solvers \cite{genie, dpm-solver, DiffVC} or by combining them with other kinds of generative models like Generative Adversarial Networks (GANs) \cite{trilemma} or Variational Autoencoders (VAEs) \cite{VDPM}. Recently proposed generative frameworks such as flow matching \cite{flow-matching} which essentially consists in optimizing the vector field corresponding to the trajectories of the probability flow Ordinary Differential Equation, or consistency models \cite{consistency-models} performing best when applied as a method of diffusion models distillation, brought performance of diffusion-related generative models quite close to that of conventional models like GANs and VAEs in terms of inference speed while preserving high quality of generated samples.

Another aspect important for any generative model is privacy preservation and robustness to various types of attacks. On the one hand, diffusion models can help other machine learning algorithms to better deal with adversarial attacks via adversarial purification \cite{densepure, adversarial-purification}. On the other hand, they tend to memorize samples from the training dataset \cite{digital-forgery} and, moreover, they are more prone to membership inference attacks than GANs with similar generation quality \cite{membership-attacks} meaning that diffusion models trained in a standard way are potentially less private than GANs. Furthermore, it has lately been shown that diffusion models can be seriously harmed by the backdoor attacks, i.e. training or fine-tuning data can be perturbed in such a way that the resulting model produces fine samples for almost all inputs but in a few cases for specific input its output is unexpected or arbitrarily bad \cite{backdoor1, backdoor2}. In particular, text-to-image diffusion models are known to be vulnerable to the presence of look-alikes (i.e. images with the same text captions but slightly different pixel content) in training datasets. This observation has been exploited to create algorithms such as Glaze and Nightshade that spoil samples produced by text-to-image models for specific prompts \cite{glaze, prompt-specific-poisoning-attacks}.

The mentioned attacks are natural tests of general robustness of diffusion models. In this work we also study a similar test and concentrate on the case when noise samples injected into training dataset are outliers, i.e. they belong to distributions different from the one diffusion is meant to be trained on. Although one can use any appropriate outlier detection algorithm to preprocess training dataset, it can be time- and resource-consuming, so we consider methods not requiring dataset filtering. Conventional diffusion model training consists in optimizing squared $L_{2}$ loss between score matching neural network output and true score function values. However, it is a well-known fact that Mean Square Error does not provide robust estimators and there exist alternatives that are less vulnerable to the presence of outliers, e.g. Huber loss function \cite{huber-loss}. This function can be seen as a smooth combination of $L_{2}$ loss for relatively small errors and $L_{1}$ loss for relatively large errors, thus serving as a robust alternative of Mean Square Error in various statistical tasks like location parameter estimation as in the original paper by Huber \yrcite{huber-loss}, or linear regression \cite{huber-regression}. A variant of Huber loss called pseudo-Huber (which we will also refer to as \textquote{\textit{P-Huber}}) loss function has already been employed in the context of diffusion-like models by Song and Dhariwal \yrcite{improved-consistency} who distilled diffusion models into consistency models with it. In this paper we explore the impact of pseudo-Huber loss function with time-dependent parameters on diffusion model robustness.

Our main contributions can be summarized as follows:
\begin{itemize}[noitemsep,topsep=0pt,parsep=0pt,partopsep=0pt]
\item We propose a novel technique of delta-scheduling by introducing a time-dependent delta parameter into pseudo-Huber loss;
\item We experimentally demonstrate the effectiveness of our approach across multiple datasets and modalities;
\item We make detailed studies on various possible schedules, pseudo-Huber loss parameters, corruption-percentages and the resilience factor computation.
\end{itemize}

\section{Preliminaries}
\label{sec:preliminaries}

This section contains general facts about diffusion probabilistic modeling and pseudo-Huber loss used in the consequent sections devoted to robust training of diffusion models.

\subsection{Diffusion Probabilistic Modeling}
\label{subsed:dpm}
Suppose we have a $n$-dimensional stochastic process of a diffusion type $X_{t}$ defined for $t\in[0,T]$ for a finite time horizon $T>0$ satisfying Stochastic Differential Equation (SDE) with drift coefficient $f(x,t): \mathbb{R}^{n}\times\mathbb{R}_{+}\to\mathbb{R}^{n}$ and positive diffusion coefficient $g_{t}$. Note that throughout this paper all SDEs are understood in It\^{o} sense \cite{SDE-book}. Anderson \yrcite{SDE-reverse} showed that under certain technical assumptions we can write down the reverse-time dynamics of this process given by the reverse SDE:

\begin{equation}
\label{eq:gt-SDE}
dX_{t} = \left(f(X_{t}, t)-g_{t}^{2}\nabla\log{p_{t}(X_{t})}\right)dt + g_{t}dW_{t} \ ,
\end{equation}
where $p_{t}$ is the marginal density of the original forward process at time $t$ and this SDE is to be solved backwards in time starting at time $T$ from random variable with density function $p_{T}$. Brownian motion $W_{t}$ driving this process is supposed to be a backward Brownian motion (meaning that its backward increments are independent, i.e. $W_{s} - W_{t}$ is independent of $W_{t}$ for $0\leq s<t\leq T$).

In classic diffusion models the data distribution is represented by $\Law{(X_{0})}$ and the drift and diffusion coefficients are chosen so that the data is gradually perturbed with Gaussian noise up to time $T$ where the noisy data $X_{T}$ is very close to standard normal distribution called the \textit{prior} of a diffusion model. If the score function $\nabla\log{p_{t}(x)}$ can be well approximated by a neural network $s_{\theta}(x,t)$, then sampling from the trained diffusion model can be performed by solving the reverse SDE (\ref{eq:gt-SDE}) backwards in time starting from a random sample from the prior distribution. Alternatively, one can choose to solve the probability flow Ordinary Differential Equation (ODE) \cite{Song-main}:

\begin{equation}
\label{eq:gt-ODE}
dX_{t} = \left(f(X_{t}, t)-\frac{g_{t}^{2}}{2}\nabla\log{p_{t}(X_{t})}\right)dt \ ,
\end{equation}
which in many cases requires less steps to produce samples of the same quality.

Song et al. \yrcite{DPM-ML-training} showed that under certain mild assumptions on data distribution minimizing weighted squared $L_{2}$ loss between the score function and neural network $s_{\theta}$ given by the expression
\begin{equation}
\label{eq:ml-dpm-loss}
\mathbb{E}_{X\sim\mu}\left[\int_{0}^{T}{g_{t}^{2} \Vert s_{\theta}(X_{t}, t) - \nabla\log{p_{t}(X_{t})}\Vert_{2}^{2}}dt\right] \ ,
\end{equation}
where $\mu$ is probability measure of paths $X=\{X_{t}\}_{t\in[0,T]}$, is equivalent to minimizing the Kullback-Leibler (KL) divergence between measure $\mu$ and the path measure corresponding to reverse diffusion parameterized with the network $s_{\theta}$. It was shown that this divergence is in fact an upper bound on negative log-likelihood of training data. Thus, $L_{2}$ loss is well justified from the point of view of both maximum likelihood training and optimizing KL divergence between path measures of reverse processes.

It is worth mentioning that although the score function cannot be computed analytically for real-world data distributions, we can make use of \textit{denoising score matching} and show that for every $t$ the following two optimization problems

\begin{equation}
\label{eq:dsm-ucond}
\min_{\theta}\leftarrow\mathbb{E}_{X_{t}}\Vert s_{\theta}(X_{t},t)-\nabla\log{p_{t}(X_{t})}\Vert_{2}^{2} \ ,
\end{equation}
\begin{equation}
\label{eq:dsm-cond}
\min_{\theta}\leftarrow\mathbb{E}_{X_{0}}\mathbb{E}_{X_{t}|X_{0}}\Vert s_{\theta}(X_{t},t)-\nabla\log{p_{t|0}(X_{t}|X_{0})}\Vert_{2}^{2}
\end{equation}
are equivalent. Note that $L_{2}$ loss is essential for the equivalence of \ref{eq:dsm-ucond} and \ref{eq:dsm-cond}. Unlike the unconditional score function $\nabla\log{p_{t}(x_{t})}$, the conditional score function $\nabla\log{p_{t|0}(x_{t}|x_{0})}$ is tractable since the conditional distribution $\Law{(X_{t}|X_{0})}$ has Gaussian densities $p_{t|0}$. Thus, training a diffusion model consists in optimizing the following squared $L_{2}$ loss

\begin{equation}
\label{eq:dpm-loss}
\mathcal{L}_{t}(X_{0})=\mathbb{E}_{X_{t}|X_{0}}\Vert s_{\theta}(X_{t},t)-\nabla\log{p_{t|0}(X_{t}|X_{0})}\Vert_{2}^{2} \ ,
\end{equation}
where $X_{0}$ is uniformly sampled from the training dataset and $t$ --- uniformly from $[0,T]$.

The framework described above stays the same for latent diffusion models \cite{VDPM}, a faster alternative of classic diffusion models, with the exception that the stochastic process $X_{t}$ is defined in latent space and $X_{0}$ corresponds to some latent data representation.

\subsection{Huber loss function}
\label{sec:huber}
One-dimensional Huber loss \cite{huber-loss} is defined as
\begin{equation}
\label{eq:huber-loss-single}
    h_{\delta}(x) =
        \begin{cases}
            \frac{1}{2}x^{2} \qquad \qquad \ \text{for} \ \ |x|\leq\delta\\
            \delta(|x| - \frac{1}{2}\delta) \ \ \ \ \ \text{for} \ \ |x|>\delta
        \end{cases}
\end{equation}
for a positive $\delta$. Its multi-dimensional version is just a coordinate-wise sum of losses $h_{\delta_{j}}(x^{j})$ of the corresponding components $x^{j}$ of the input vector $x\in\mathbb{R}^{n}$. Parameters $\delta_{j}$ for $j=1,..,n$ can be different. When computed on the difference between the true value and its prediction by some statistical model, this function penalizes the small errors like Mean Square Error (MSE) loss and the large errors like Mean Absolute Error (MAE) loss. Thus, Huber loss penalizes the large errors caused by outliers less than MSE loss which makes it attractive for robust statistical methods \cite{huber-regression}.

It is easy to see that derivative of Huber loss is continuous, but not differentiable in points $|x|=\delta$. Pseudo-Huber loss is a more smooth function also behaving like MSE in one-dimensional case in the neighbourhood of zero and like MAE in the neighbourhood of infinity:

\begin{equation}
\label{eq:pseudo-huber-loss}
H_{\delta}(x) = \delta^{2}\left(\sqrt{1+\frac{x^{2}}{\delta^{2}}} - 1\right) \ .
\end{equation}
This function is defined for positive values of the parameter $\delta$ controlling its tolerance to errors with a large $L_{2}$ norm. It is extended to multi-dimensional input in the coordinate-wise manner as the standard Huber loss $h_{\delta}(x)$.

In this paper we study the following training objective for diffusion models expressed in terms of pseudo-Huber loss:
\begin{equation}
\label{eq:dpm-h-loss}
\mathcal{L}^{(\delta)}_{t}(X_{0})=\mathbb{E}_{X_{t}|X_{0}}H_{\delta}\left( s_{\theta}^{(\delta)}(X_{t},t)-\nabla\log{p_{t|0}(X_{t}|X_{0})}\right) ,
\end{equation}
where $\delta$ can be time-dependent.

\section{Experiments}
\label{sec:experiments}

\subsection{Text-to-Image}
\label{subsed:t2i}

\begin{figure*}[ht]
\vskip 0.05in
\begin{center}
\includegraphics[width=1.0\linewidth]{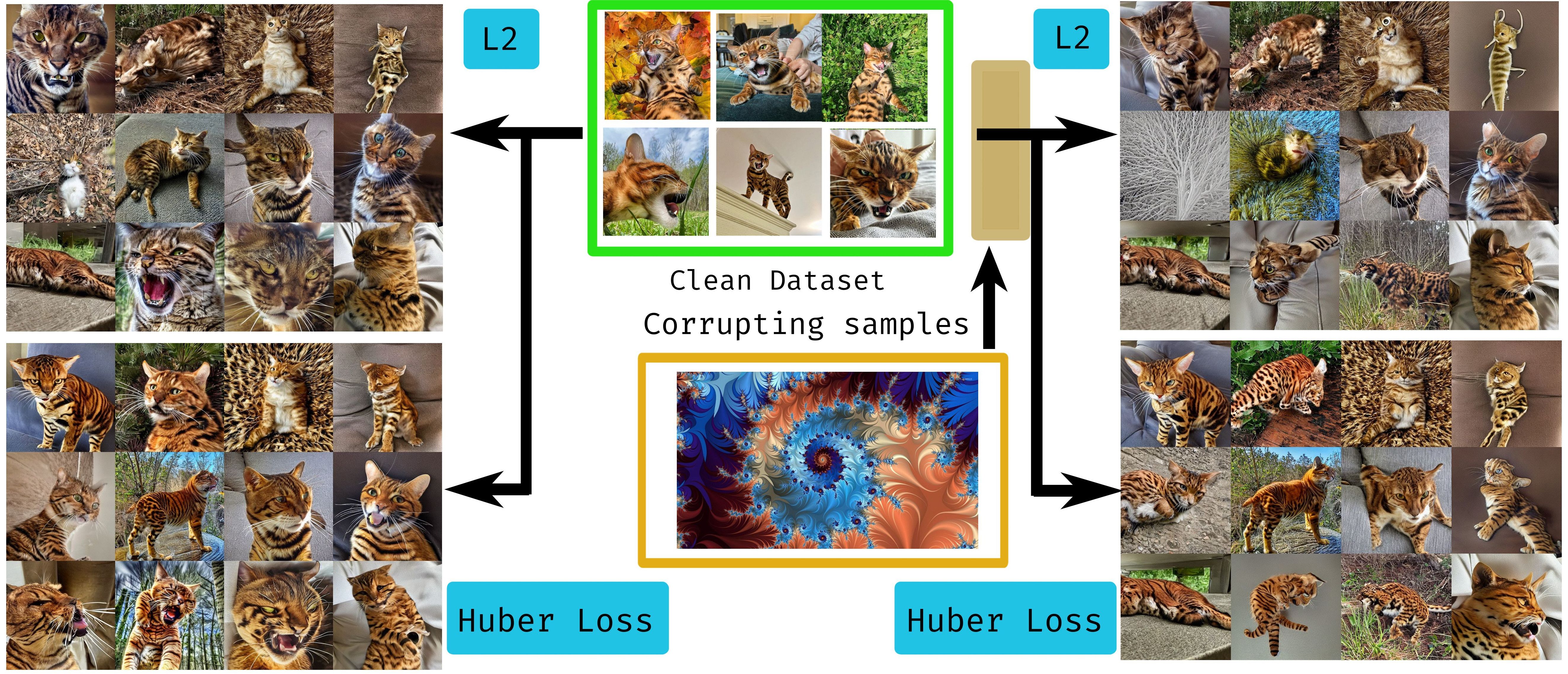}
\caption{Scheme of the process. Off-topic images are added to a clean dataset of cat photos. When the L2 loss function is used, it leads to concept distortion and even erasure (see the fractal-like structures in the second row). Meanwhile the Huber loss training results stay consistent with their not-corrupted counterparts.}
\label{fig:example}
\end{center}
\vskip -0.05in
\end{figure*}

For text2image customization experiments we used the Dreambooth framework \cite{ruiz2023dreambooth} on Stable Diffusion v1.5 implemented in Huggingface Diffusers \cite{diffusers} library. We used LoRA \cite{hu2021lora} technique for faster training. We tested adaptation on $7$ different datasets in various domains: characters, styles and landscapes.

We tested $4$ levels of corruption: $0\% (\text{clean run}), 15\%, 30\%, 45\%$. For each level of corruption a corresponding portion of images from the target dataset ("clean") was replaced with samples from the other datasets having the lowest cosine similarity of CLIP embeddings \cite{clip} with random "clean" samples. For each experiment 3 models have been trained: the model with $L_{2}$ loss, the model with pseudo-Huber loss and the model with pseudo-Huber loss delta scheduling for which the $\delta$ parameter depended on the time step $t$. During the preliminary experiments we chose the exponential decrease scheduler for $\delta$. Other schedules are possible, but not all of them are optimal. Most importantly, the schedules with increasing $\delta$ have yielded the worst results, supporting the hypothesis that it's much more natural for the parameter to deplete over timesteps and ensure well image approximation at the later steps of the reverse-diffusion process. Appendix \ref{app:schedules} for the ablation studies on different other possible schedules.

We sampled images from the trained models and for each \textit{reference}-\textit{sample} pair -- consisting of a reference image from "clean" or "corrupting" datasets and a sampled image -- we computed CLIP similarity and \textquote{$1 - \text{LPIPS}$} score \cite{lpips}, then calculated the mean across all pairs.

Now we had the average similarity scores to the clean data reference images:

\begin{itemize}[noitemsep,topsep=0pt,parsep=0pt,partopsep=0pt]
    \item when the dataset had been corrupted $S_{\text{to clean}}^{\text{corrupted}}$;
    \item when it had been trained on the clean dataset $S_{\text{to clean}}^{\text{clean}}$.
\end{itemize}

Through the selected similarity metric of these we introduce the key $R$-variable representing how well the model fares against corruption:

\begin{equation}
\label{resilience}
R = S_{\text{to clean}}^{\text{corrupted}}-S_{\text{to clean}}^{\text{clean}}
\end{equation}

We used similarity difference instead of, for example, division, as it's more stable in regards to zeroing of one of its components and the event of their sign change. See Appendix \ref{app:r-factor-comparison-plots} for the numerical and visual comparison of the results obtained for the different versions of the $R$-value and the detailed reasoning behind using the differential metric.

CLIP-derived metrics have been shown to be inaccurate to measure image features similarity, especially in the context of deflecting adversarial attacks.\cite{shan2023response} Because of it, and that LPIPS has been giving more consistent results than CLIP, we decided to use the \textquote{1 - LPIPS} similarity metric for most of our work. We provide the CLIP and LPIPS comparison plots and show their qualitatively close and different parts in Appendix \ref{app:clip-lpips-comparison}.

\begin{figure}[h]
\begin{center}
\includegraphics[width=0.98\linewidth]{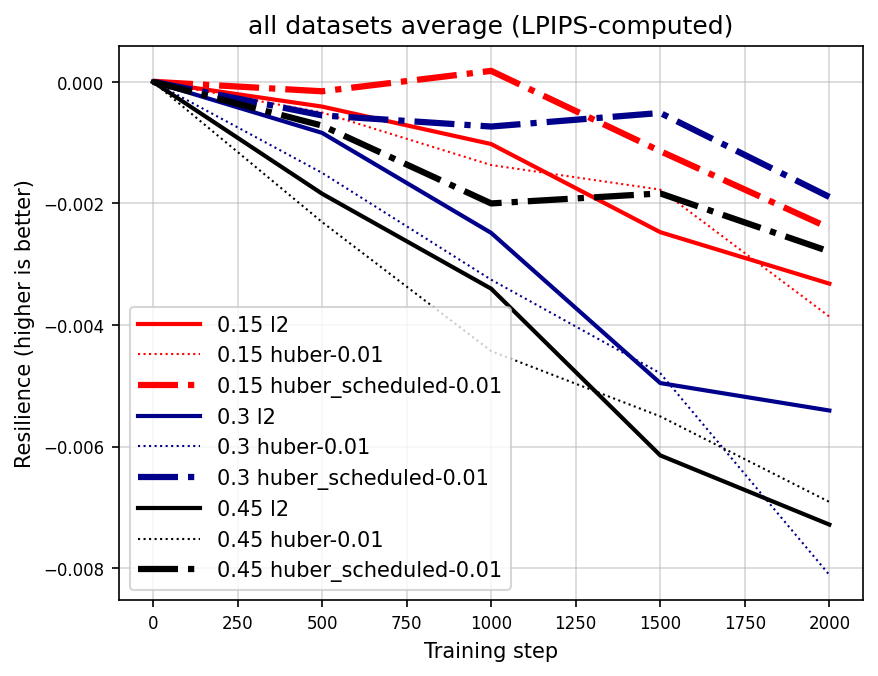}
\caption{The plot of the Resilience factor for \textquote{1 - LPIPS} similarity for all the tested text2image prompts at different levels of corruption at the selected $\delta = 0.01$.}
\label{fig:all-prompts-study}
\end{center}
\end{figure}

\begin{table*}[h]
\begin{tabular}{lrrrr}
\toprule
 & a landscape of sks & a photo of a shoan & a photo of an eichpoch & a picture by david revoy \\
\midrule
l2 & -0.38 & \textbf{-0.01} & \textbf{0.00} & -0.08 \\
huber-0.1 & -0.28 & -0.20 & -0.02 & -0.09 \\
huber scheduled-0.1 & \textbf{-0.06} & -0.13 & -0.22 & \textbf{-0.03} \\
\midrule
& lordbob\footnote{a picture in the style of lordbob} & a picture of a drake & maxwell the cat & --- \\
\midrule
l2 & -0.42 & -0.39 & -0.14 & --- \\
huber-0.1 & -0.33 & -0.40 & -0.13 & --- \\
huber scheduled-0.1 & \textbf{-0.11} & \textbf{-0.14} & \textbf{-0.04} & --- \\
\bottomrule
\end{tabular}
\caption{R-scores for all the tested text2image prompts/concepts at 0.45 of corruption, 0.1 Huber $\delta_0$ and 2000 steps. To take a look at examples of each concept, see Appendix \ref{app:datasets}. To get an overview of how these functions change over the model's training steps, see the plots in Appendix \ref{app:prompts-comparison-plots}.}
\label{table:prompt-comparison}
\end{table*}

The comparison of Scheduled Pseudo-Huber, Huber and $L_{2}$ losses in terms of $R$-scores on all datasets at the final training step is listed at Table \ref{table:prompt-comparison}. It is evident that Scheduled Pseudo-Huber loss outperforms $L_2$ in 5 out of 7 cases although the degree of its advantage varies. To see the training-wide dynamics from step 0 to 2000 please refer to Appendix \ref{app:prompts-comparison-plots} where it's shown that this advantage is largely consistent on all later training steps.

\subsection{Text-to-Speech}
\label{subsed:t2s}

For speech domain few-shot speaker adaptation scenario was chosen. We fine-tuned a pre-trained multi-speaker Grad-TTS \cite{Grad-TTS} model decoder for a new voice. The experiment was conducted for $10$ female and $4$ male speakers. Three datasets for each speaker were constructed: "clean" dataset with $16$ records of the target speaker and $2$ "corrupted" datasets with additional $4$ "corrupting" records by a female and a male voice correspondingly. 

We tested performance of Grad-TTS models fined-tuned on corrupted datasets with $L_2$ and pseudo-Huber loss. We also fine-tuned Grad-TTS with $L_2$ loss on clean data. As in the previous case, we used pseudo-Huber loss with exponential decrease scheduler in our experiments.

As a target metric, we used speaker similarity of synthesized speech evaluated with a pre-trained speaker verification model\footnote{https://github.com/CorentinJ/Real-Time-Voice-Cloning}. 

\begin{figure}[h]
\begin{center}
\includegraphics[width=0.98\linewidth]{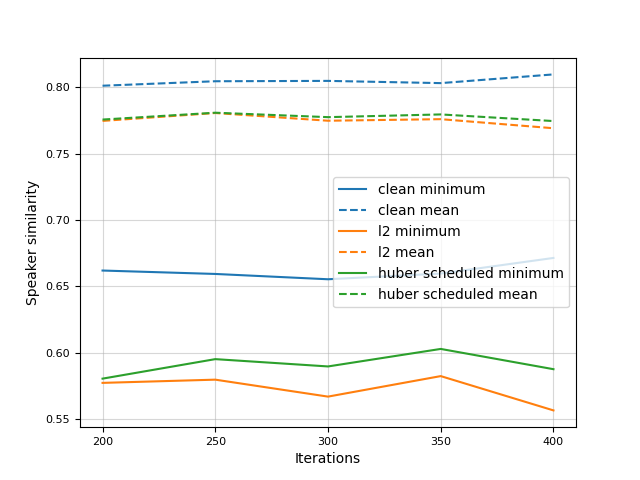}
\caption{Speaker similarity for different iterations averaged across speakers. Models: \textit{clean} - trained on clean dataset; \textit{$l_2$} and \textit{huber scheduled} - trained on mixed dataset with corresponding losses.}
\label{fig:speech_similarity}
\end{center}
\end{figure}

\begin{figure}[h]
\begin{center}
\includegraphics[width=0.98\linewidth]{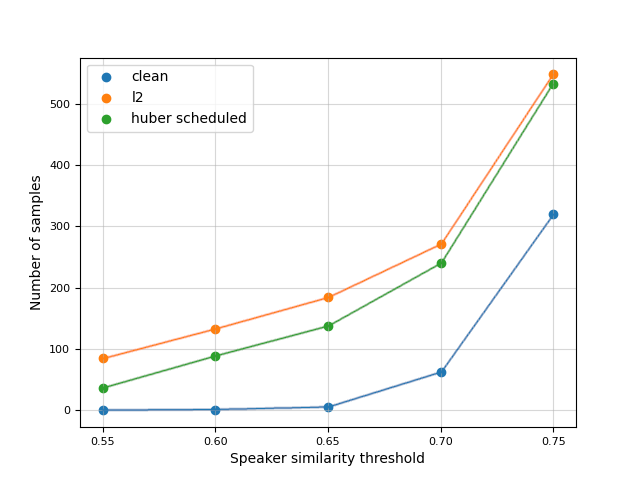}
\caption{Number of synthesized samples with similarity less than corresponding threshold. Total number of samples $1260$ from the best checkpoints on $350$ iterations.}
\label{fig:speech_count}
\end{center}
\end{figure}

We generated $90$ samples for every speaker and every model and calculated mean and minimum similarity values. We calculated minimum similarity by averaging similarity score between a generated sample and $5$ clean samples. Then the minimum value across generated samples was chosen. In case of mean similarity the mean value across generated samples was calculated.

According to Figure \ref{fig:speech_similarity} models trained on corrupted datasets show quite high mean similarity despite a slight drop compared to the model trained on clean data. Although we observe a comparable drop in minimum similarity, models with pseudo-Huber loss still constantly demonstrate better score on all iterations in terms of minimum similarity. Furthermore, we analysed samples from the best checkpoint ($350$ iterations) depending on different similarity thresholds. Figure \ref{fig:speech_count} illustrates that standard $L_2$ training scheme leads to samples with lower similarity more often which means that training with Huber loss with scheduling is more robust than standard $L_2$-based approach.

\section{Limitations and Future Work}
\label{sec:limits}

While protecting the models as a whole from some parts of the dataset being poisoned, this method will most likely not work in case the \textit{entire} dataset has been corrupted through some sort of an adversarial attack -- meaning the big players will be fine, while the small models creators might still be vulnerable to style-transfer failure. The conditions over the data structure, which may be putting limits upon the data-corruption techniques, is still unknown, however the knowledge of Scheduled Pseudo-Huber loss can spawn a new generation of such algorithms. 
Our work doesn't include tests against the SOTA poisoning models because it will look like active data-protection removal, and we will leave it to the community.

\section{Conclusion}
\label{sec:conclusion}
In this paper we investigated the possibility of using Huber loss for training Diffusion Probabilistic models. We presented the sufficient conditions for the utility of Huber loss when dealing with dataset corruption. Moreover, our theoretical analysis predicts that Huber loss might require different delta parameters at different time steps of the backward diffusion process. We provided experimental evidence of better performance of Huber loss and Huber loss with time dependent delta parameter in model adaptation setting for both image and audio domain. Further research in this area may involve identifying more precise conditions for the applicability of Huber loss for robust model adaptation. Additionally, a deeper analysis of scheduling of the delta parameter could be an important research direction.

\section{Ethical Statement}
\label{sec:ethical}
Because this data-corruption resilience scheme has virtually no cost over $L_2$ loss computation, it will save considerably more resources -- that can go to more society-beneficial goals -- as compared to using neural networks to re-caption, filter out or "purify" the images, saving money and time for large-scale model trainers and helping to preserve the environment. 

\bibliography{huber_paper}

\begin{thebibliography}{40}
\providecommand{\natexlab}[1]{#1}
\providecommand{\url}[1]{\texttt{#1}}
\expandafter\ifx\csname urlstyle\endcsname\relax
  \providecommand{\doi}[1]{doi: #1}\else
  \providecommand{\doi}{doi: \begingroup \urlstyle{rm}\Url}\fi

\bibitem[Anderson(1982)]{SDE-reverse}
Anderson, B.~D.
\newblock {Reverse-time Diffusion Equation Models}.
\newblock \emph{Stochastic Processes and their Applications}, 12\penalty0 (3):\penalty0 313 -- 326, 1982.
\newblock ISSN 0304-4149.

\bibitem[Carlini et~al.(2023)Carlini, Hayes, Nasr, Jagielski, Sehwag, Tram{\`e}r, Balle, Ippolito, and Wallace]{membership-attacks}
Carlini, N., Hayes, J., Nasr, M., Jagielski, M., Sehwag, V., Tram{\`e}r, F., Balle, B., Ippolito, D., and Wallace, E.
\newblock {Extracting Training Data from Diffusion Models}.
\newblock In \emph{32nd USENIX Security Symposium (USENIX Security 23)}, pp.\  5253--5270. USENIX Association, aug 2023.

\bibitem[Das et~al.(2023)Das, Yang, Hospedales, Xiang, and Song]{chirodiff}
Das, A., Yang, Y., Hospedales, T., Xiang, T., and Song, Y.-Z.
\newblock {ChiroDiff: Modelling chirographic data with Diffusion Models}.
\newblock In \emph{The Eleventh International Conference on Learning Representations}, 2023.

\bibitem[Dhariwal \& Nichol(2021)Dhariwal and Nichol]{DPMs-beat-GANs}
Dhariwal, P. and Nichol, A.
\newblock {Diffusion Models Beat GANs on Image Synthesis}.
\newblock In \emph{Advances in Neural Information Processing Systems}, volume~34, pp.\  8780--8794. Curran Associates, Inc., 2021.

\bibitem[Dockhorn et~al.(2022)Dockhorn, Vahdat, and Kreis]{genie}
Dockhorn, T., Vahdat, A., and Kreis, K.
\newblock {GENIE: Higher-Order Denoising Diffusion Solvers}.
\newblock In \emph{Advances in Neural Information Processing Systems}, volume~35, pp.\  30150--30166. Curran Associates, Inc., 2022.

\bibitem[Gao et~al.(2023)Gao, Liu, Zeng, Xu, Li, Luo, Liu, Zhen, and Zhang]{image-superres}
Gao, S., Liu, X., Zeng, B., Xu, S., Li, Y., Luo, X., Liu, J., Zhen, X., and Zhang, B.
\newblock {Implicit Diffusion Models for Continuous Super-Resolution}.
\newblock In \emph{Proceedings of the IEEE/CVF Conference on Computer Vision and Pattern Recognition (CVPR)}, pp.\  10021--10030, June 2023.

\bibitem[Han \& Lee(2022)Han and Lee]{nuwave2}
Han, S. and Lee, J.
\newblock {NU-Wave 2: A General Neural Audio Upsampling Model for Various Sampling Rates}.
\newblock In \emph{Proc. Interspeech 2022}, pp.\  4401--4405, 2022.

\bibitem[Hawthorne et~al.(2022)Hawthorne, Simon, Roberts, Zeghidour, Gardner, Manilow, and Engel]{music-generation}
Hawthorne, C., Simon, I., Roberts, A., Zeghidour, N., Gardner, J., Manilow, E., and Engel, J.~H.
\newblock {Multi-instrument Music Synthesis with Spectrogram Diffusion}.
\newblock In \emph{Proceedings of the 23rd International Society for Music Information Retrieval Conference, {ISMIR} 2022, Bengaluru, India, December 4-8, 2022}, pp.\  598--607, 2022.

\bibitem[Ho et~al.(2020)Ho, Jain, and Abbeel]{DDPM}
Ho, J., Jain, A., and Abbeel, P.
\newblock {Denoising Diffusion Probabilistic Models}.
\newblock In \emph{Advances in Neural Information Processing Systems 33: Annual Conference on Neural Information Processing Systems 2020, NeurIPS 2020, December 6-12, 2020, virtual}, volume~33. Curran Associates, Inc., 2020.

\bibitem[Hu et~al.(2021)Hu, Shen, Wallis, Allen-Zhu, Li, Wang, Wang, and Chen]{hu2021lora}
Hu, E.~J., Shen, Y., Wallis, P., Allen-Zhu, Z., Li, Y., Wang, S., Wang, L., and Chen, W.
\newblock Lora: Low-rank adaptation of large language models, 2021.

\bibitem[Huber(1964)]{huber-loss}
Huber, P.~J.
\newblock {Robust Estimation of a Location Parameter}.
\newblock \emph{The Annals of Mathematical Statistics}, 35\penalty0 (1):\penalty0 73 -- 101, 1964.

\bibitem[Kawar et~al.(2022)Kawar, Elad, Ermon, and Song]{DDRM}
Kawar, B., Elad, M., Ermon, S., and Song, J.
\newblock {Denoising Diffusion Restoration Models}.
\newblock In \emph{Advances in Neural Information Processing Systems}, volume~35, pp.\  23593--23606. Curran Associates, Inc., 2022.

\bibitem[Kingma et~al.(2021)Kingma, Salimans, Poole, and Ho]{VDPM}
Kingma, D., Salimans, T., Poole, B., and Ho, J.
\newblock {Variational Diffusion Models}.
\newblock In \emph{Advances in Neural Information Processing Systems}, volume~34, pp.\  21696--21707. Curran Associates, Inc., 2021.

\bibitem[Lipman et~al.(2023)Lipman, Chen, Ben-Hamu, Nickel, and Le]{flow-matching}
Lipman, Y., Chen, R. T.~Q., Ben-Hamu, H., Nickel, M., and Le, M.
\newblock {Flow Matching for Generative Modeling}.
\newblock In \emph{The Eleventh International Conference on Learning Representations}, 2023.

\bibitem[Liptser \& Shiryaev(1978)Liptser and Shiryaev]{SDE-book}
Liptser, R.~S. and Shiryaev, A.~N.
\newblock \emph{{Statistics of Random Processes}}, volume~5 of \emph{Stochastic Modelling and Applied Probability}.
\newblock Springer-Verlag, 1978.

\bibitem[Lu et~al.(2022)Lu, Zhou, Bao, Chen, LI, and Zhu]{dpm-solver}
Lu, C., Zhou, Y., Bao, F., Chen, J., LI, C., and Zhu, J.
\newblock {DPM-Solver: A Fast ODE Solver for Diffusion Probabilistic Model Sampling in Around 10 Steps}.
\newblock In Koyejo, S., Mohamed, S., Agarwal, A., Belgrave, D., Cho, K., and Oh, A. (eds.), \emph{Advances in Neural Information Processing Systems}, volume~35, pp.\  5775--5787. Curran Associates, Inc., 2022.

\bibitem[Luo et~al.(2023)Luo, Chen, Zhang, Huang, Wang, Shen, Zhao, Zhou, and Tan]{videofusion}
Luo, Z., Chen, D., Zhang, Y., Huang, Y., Wang, L., Shen, Y., Zhao, D., Zhou, J., and Tan, T.
\newblock {VideoFusion: Decomposed Diffusion Models for High-Quality Video Generation}.
\newblock In \emph{2023 IEEE/CVF Conference on Computer Vision and Pattern Recognition (CVPR)}, pp.\  10209--10218. IEEE Computer Society, jun 2023.

\bibitem[Meng et~al.(2022)Meng, He, Song, Song, Wu, Zhu, and Ermon]{sdedit}
Meng, C., He, Y., Song, Y., Song, J., Wu, J., Zhu, J.-Y., and Ermon, S.
\newblock {SDEdit: Guided Image Synthesis and Editing with Stochastic Differential Equations}.
\newblock In \emph{International Conference on Learning Representations}, 2022.

\bibitem[Nie et~al.(2022)Nie, Guo, Huang, Xiao, Vahdat, and Anandkumar]{adversarial-purification}
Nie, W., Guo, B., Huang, Y., Xiao, C., Vahdat, A., and Anandkumar, A.
\newblock {Diffusion Models for Adversarial Purification}.
\newblock In \emph{Proceedings of the 39th International Conference on Machine Learning}, volume 162 of \emph{Proceedings of Machine Learning Research}, pp.\  16805--16827. PMLR, 17--23 Jul 2022.

\bibitem[Owen(2007)]{huber-regression}
Owen, A.~B.
\newblock {A robust hybrid of lasso and ridge regression}.
\newblock \emph{Contemporary Mathematics}, 443:\penalty0 59 -- 72, 01 2007.

\bibitem[Popov et~al.(2021)Popov, Vovk, Gogoryan, Sadekova, and Kudinov]{Grad-TTS}
Popov, V., Vovk, I., Gogoryan, V., Sadekova, T., and Kudinov, M.
\newblock {Grad-TTS: A Diffusion Probabilistic Model for Text-to-Speech}.
\newblock In \emph{Proceedings of the 38th International Conference on Machine Learning, {ICML} 2021, 18-24 July 2021, Virtual Event}, volume 139 of \emph{Proceedings of Machine Learning Research}, pp.\  8599--8608. {PMLR}, 2021.

\bibitem[Popov et~al.(2022)Popov, Vovk, Gogoryan, Sadekova, Kudinov, and Wei]{DiffVC}
Popov, V., Vovk, I., Gogoryan, V., Sadekova, T., Kudinov, M., and Wei, J.
\newblock {Diffusion-Based Voice Conversion with Fast Maximum Likelihood Sampling Scheme}.
\newblock In \emph{International Conference on Learning Representations}, 2022.

\bibitem[Radford et~al.(2021)Radford, Kim, Hallacy, Ramesh, Goh, Agarwal, Sastry, Askell, Mishkin, Clark, Krueger, and Sutskever]{clip}
Radford, A., Kim, J.~W., Hallacy, C., Ramesh, A., Goh, G., Agarwal, S., Sastry, G., Askell, A., Mishkin, P., Clark, J., Krueger, G., and Sutskever, I.
\newblock Learning transferable visual models from natural language supervision, 2021.

\bibitem[Rombach et~al.(2022)Rombach, Blattmann, Lorenz, Esser, and Ommer]{stable-diffusion}
Rombach, R., Blattmann, A., Lorenz, D., Esser, P., and Ommer, B.
\newblock {High-Resolution Image Synthesis With Latent Diffusion Models}.
\newblock In \emph{Proceedings of the IEEE/CVF Conference on Computer Vision and Pattern Recognition (CVPR)}, pp.\  10684--10695, June 2022.

\bibitem[Ruiz et~al.(2023)Ruiz, Li, Jampani, Pritch, Rubinstein, and Aberman]{ruiz2023dreambooth}
Ruiz, N., Li, Y., Jampani, V., Pritch, Y., Rubinstein, M., and Aberman, K.
\newblock Dreambooth: Fine tuning text-to-image diffusion models for subject-driven generation, 2023.

\bibitem[Shan et~al.(2023{\natexlab{a}})Shan, Cryan, Wenger, Zheng, Hanocka, and Zhao]{glaze}
Shan, S., Cryan, J., Wenger, E., Zheng, H., Hanocka, R., and Zhao, B.~Y.
\newblock {Glaze: Protecting Artists from Style Mimicry by Text-to-Image Models}.
\newblock In \emph{Proceedings of the 32nd USENIX Conference on Security Symposium}, SEC '23. USENIX Association, 2023{\natexlab{a}}.

\bibitem[Shan et~al.(2023{\natexlab{b}})Shan, Ding, Passananti, Zheng, and Zhao]{prompt-specific-poisoning-attacks}
Shan, S., Ding, W., Passananti, J., Zheng, H., and Zhao, B.~Y.
\newblock {Prompt-Specific Poisoning Attacks on Text-to-Image Generative Models}, 2023{\natexlab{b}}.

\bibitem[Shan et~al.(2023{\natexlab{c}})Shan, Wu, Zheng, and Zhao]{shan2023response}
Shan, S., Wu, S., Zheng, H., and Zhao, B.~Y.
\newblock A response to glaze purification via impress, 2023{\natexlab{c}}.

\bibitem[Somepalli et~al.(2023)Somepalli, Singla, Goldblum, Geiping, and Goldstein]{digital-forgery}
Somepalli, G., Singla, V., Goldblum, M., Geiping, J., and Goldstein, T.
\newblock {Diffusion Art or Digital Forgery? Investigating Data Replication in Diffusion Models}.
\newblock In \emph{2023 IEEE/CVF Conference on Computer Vision and Pattern Recognition (CVPR)}, pp.\  6048--6058. IEEE Computer Society, jun 2023.

\bibitem[Song \& Dhariwal(2023)Song and Dhariwal]{improved-consistency}
Song, Y. and Dhariwal, P.
\newblock {Improved Techniques for Training Consistency Models}, 2023.

\bibitem[Song et~al.(2021{\natexlab{a}})Song, Durkan, Murray, and Ermon]{DPM-ML-training}
Song, Y., Durkan, C., Murray, I., and Ermon, S.
\newblock {Maximum Likelihood Training of Score-Based Diffusion Models}.
\newblock In \emph{Advances in Neural Information Processing Systems}, volume~34, pp.\  1415--1428. Curran Associates, Inc., 2021{\natexlab{a}}.

\bibitem[Song et~al.(2021{\natexlab{b}})Song, Sohl-Dickstein, Kingma, Kumar, Ermon, and Poole]{Song-main}
Song, Y., Sohl-Dickstein, J., Kingma, D.~P., Kumar, A., Ermon, S., and Poole, B.
\newblock {Score-Based Generative Modeling through Stochastic Differential Equations}.
\newblock In \emph{International Conference on Learning Representations}, 2021{\natexlab{b}}.

\bibitem[Song et~al.(2023)Song, Dhariwal, Chen, and Sutskever]{consistency-models}
Song, Y., Dhariwal, P., Chen, M., and Sutskever, I.
\newblock {Consistency models}.
\newblock In \emph{Proceedings of the 40th International Conference on Machine Learning}, ICML'23. JMLR.org, 2023.

\bibitem[Struppek et~al.(2023)Struppek, Hentschel, Poth, Hintersdorf, and Kersting]{backdoor2}
Struppek, L., Hentschel, M., Poth, C., Hintersdorf, D., and Kersting, K.
\newblock {Leveraging Diffusion-Based Image Variations for Robust Training on Poisoned Data}.
\newblock In \emph{NeurIPS 2023 Workshop on Backdoors in Deep Learning - The Good, the Bad, and the Ugly}, 2023.

\bibitem[Tevet et~al.(2023)Tevet, Raab, Gordon, Shafir, Cohen-or, and Bermano]{human-motion}
Tevet, G., Raab, S., Gordon, B., Shafir, Y., Cohen-or, D., and Bermano, A.~H.
\newblock {Human Motion Diffusion Model}.
\newblock In \emph{The Eleventh International Conference on Learning Representations}, 2023.

\bibitem[von Platen et~al.(2022)von Platen, Patil, Lozhkov, Cuenca, Lambert, Rasul, Davaadorj, and Wolf]{diffusers}
von Platen, P., Patil, S., Lozhkov, A., Cuenca, P., Lambert, N., Rasul, K., Davaadorj, M., and Wolf, T.
\newblock {Diffusers: State-of-the-Art Diffusion Models}.
\newblock \url{https://github.com/huggingface/diffusers}, 2022.

\bibitem[Wang et~al.(2023)Wang, Shen, Tong, Zhang, and Kawaguchi]{backdoor1}
Wang, H., Shen, Q., Tong, Y., Zhang, Y., and Kawaguchi, K.
\newblock {The Stronger the Diffusion Model, the Easier the Backdoor: Data Poisoning to Induce Copyright Breaches Without Adjusting Finetuning Pipeline}.
\newblock In \emph{NeurIPS 2023 Workshop on Backdoors in Deep Learning - The Good, the Bad, and the Ugly}, 2023.

\bibitem[Xiao et~al.(2023)Xiao, Chen, Jin, Wang, Nie, Liu, Anandkumar, Li, and Song]{densepure}
Xiao, C., Chen, Z., Jin, K., Wang, J., Nie, W., Liu, M., Anandkumar, A., Li, B., and Song, D.
\newblock {DensePure: Understanding Diffusion Models for Adversarial Robustness}.
\newblock In \emph{The Eleventh International Conference on Learning Representations}, 2023.

\bibitem[Xiao et~al.(2022)Xiao, Kreis, and Vahdat]{trilemma}
Xiao, Z., Kreis, K., and Vahdat, A.
\newblock {Tackling the Generative Learning Trilemma with Denoising Diffusion GANs}.
\newblock In \emph{International Conference on Learning Representations}, 2022.

\bibitem[Zhang et~al.(2018)Zhang, Isola, Efros, Shechtman, and Wang]{lpips}
Zhang, R., Isola, P., Efros, A.~A., Shechtman, E., and Wang, O.
\newblock The unreasonable effectiveness of deep features as a perceptual metric.
\newblock In \emph{CVPR}, 2018.

\end{thebibliography}
\bibliographystyle{icml2024}

\newpage
\appendix
\onecolumn

\section{Ablation studies}
\label{app:more-plots}

\subsection{Impact of different PHL schedules}
\label{app:schedules}

\begin{figure}[!b] 
    \centering
   \subfigure[]{
       \includegraphics[width=0.42\textwidth]{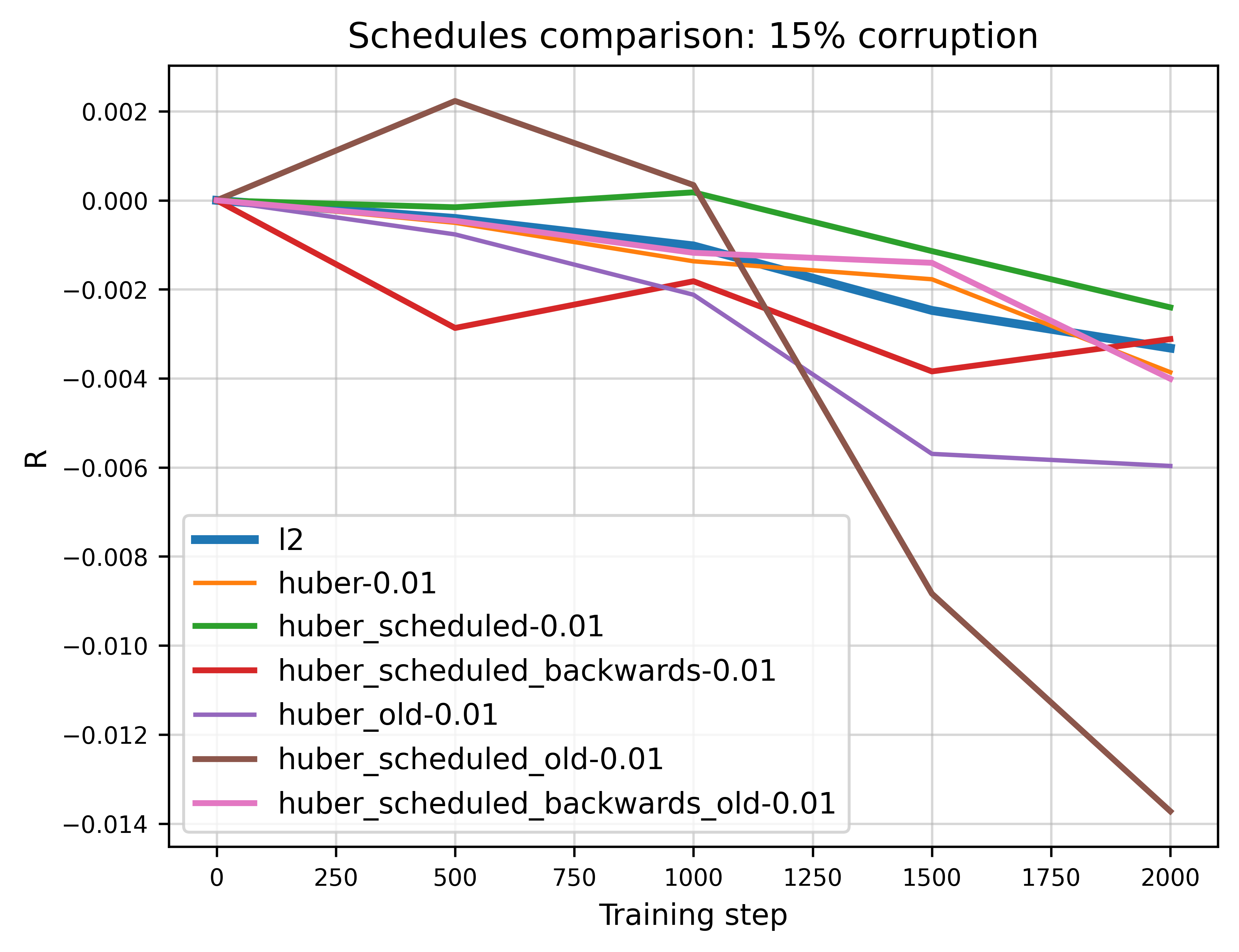}
       \label{fig:subim1}
   }
   \subfigure[]{
       \includegraphics[width=0.42\textwidth]{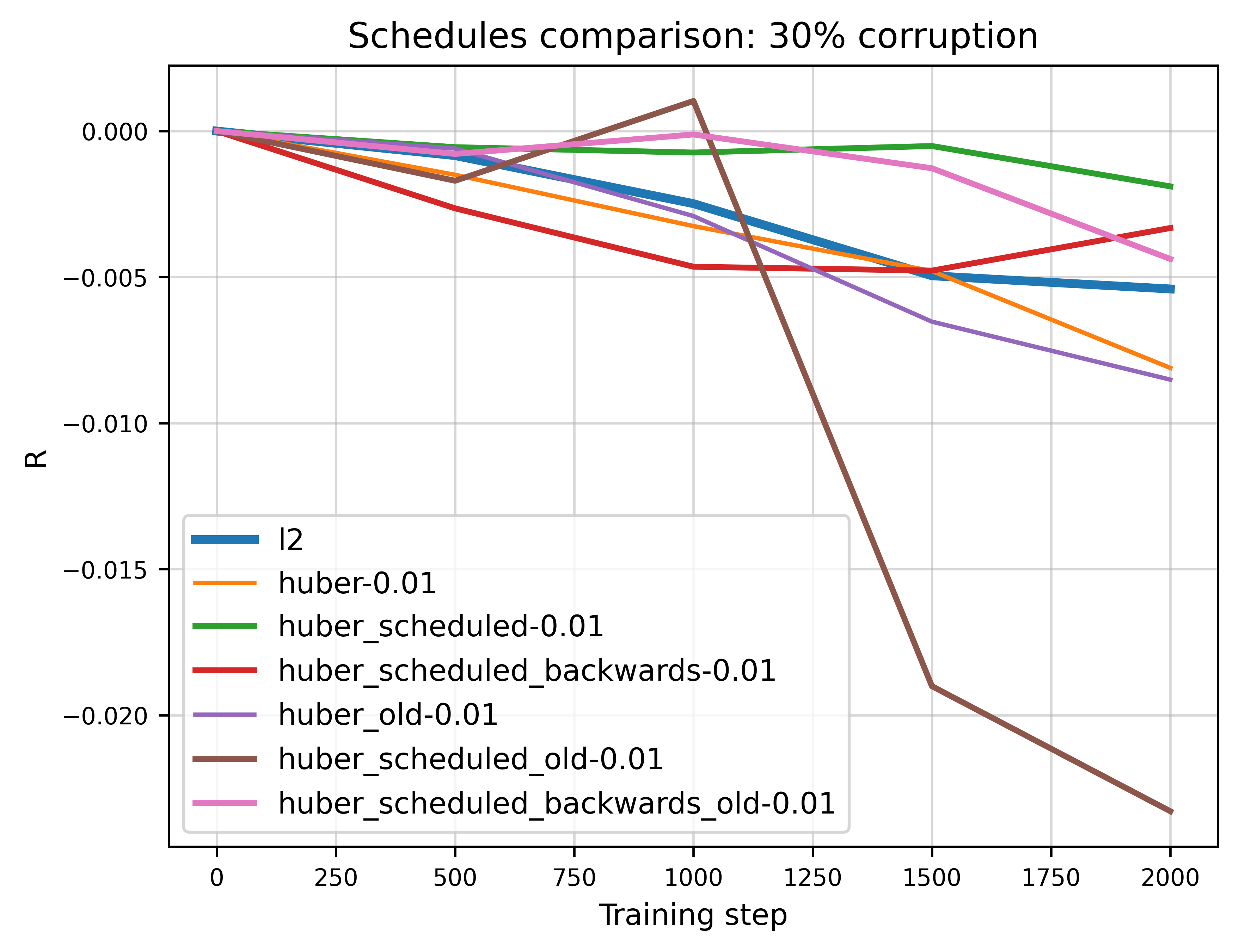}
       \label{fig:subim2}
   }
   \subfigure[]{
       \includegraphics[width=0.42\textwidth]{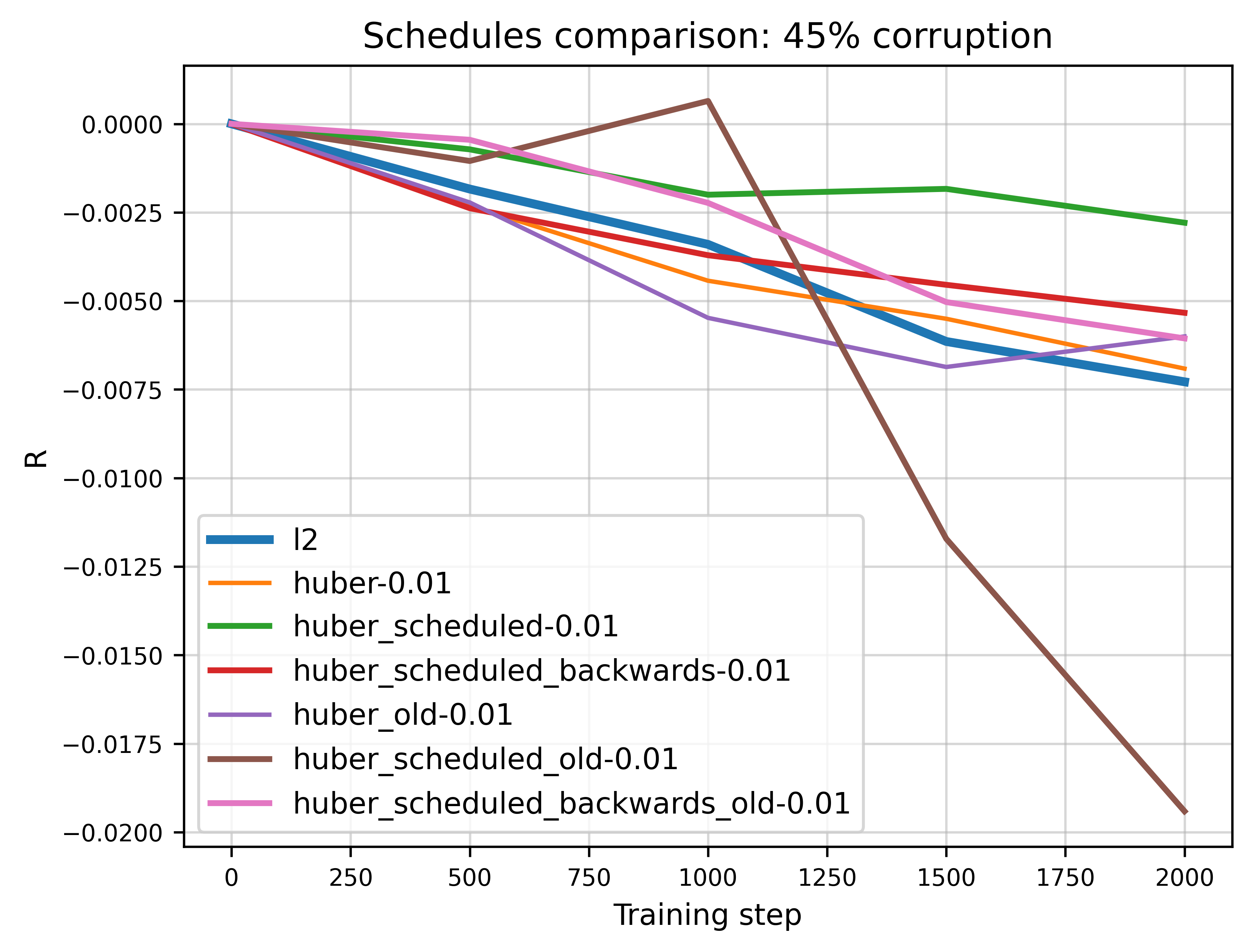}
       \label{fig:subim3}
   }

   \caption{A study of various Pseudo-Huber loss schedulers's \textquote{LPIPS}-calculated resilience, averaged across all the prompts used and computed for different amounts of corruption. (Pseudo-)Huber losses with the postfix \textquote{old} are the implementation of (\ref{pseudo-huber-loss-diffusers}), while ones without are the \textquote{corrected} versions (\ref{eq:pseudo-huber-loss}). The schedules used are the \textquote{backwards} and the \textquote{forward} (without the note) versions of the exponential schedule (\ref{exp-schedule}); it is evident that the \textquote{backwards} version fails significantly. It's evident that our final adopted schedule outperforms L2 and all the other tested variants.}
\label{fig:schedule-study3}
\end{figure}

We adopted a simple exponential decrease/increase for the constant $\delta$ in the schedule of Pseudo-Huber loss. The scheduling for the exponential decrease is given by the formula


\begin{equation}
\label{exp-schedule}
\delta = \exp\left({\frac{\log{\delta_0} * \text{timestep}}{\text{num train timesteps}}}\right)
\end{equation}
The exponential increase ("backwards") is obtained by time reversal $[\text{timestep} := \text{num train timesteps} - \text{timestep}]$: 



In the process of our experiments it turned out the Diffusers implementation of pseudo-Huber loss

\begin{equation}
\label{pseudo-huber-loss-diffusers}
H_{c}^{\text{diffusers}}(x) = \sqrt{|x|^{2} + c^{2}} - c \ ,
\end{equation}
with the same coordinate-wise extension to multi-dimensional case as losses (\ref{eq:huber-loss-single}) and (\ref{eq:pseudo-huber-loss}), was not fully mathematically correct as it lacked the leading $c$ coefficient resulting in wrong asymptotics for large values of parameter ($H_{c}^{diffusers} \sim \frac{1}{2}\frac{|x|^{2}}{c}$ for large values of $c$ instead of correct asymptotics $\frac{|x|^{2}}{2}$ of pseudo-Huber loss $H_{\delta}$). While it didn't influence normal Huber runs because the loss was proportional to the true one, it was prominent in our cases of time-dependent $\delta$. Though the formula \ref{pseudo-huber-loss-diffusers} is incorrect we decided to include schedulers based in it into comparison to demonstrate its drastically inferior performance caused by the incorrect asymptotics.


To decide what schedule is better and to test the claim of the theorem about the limiting cases values (that were overshadowed by the loss's secondary dependence on $c$), we include the full comparison here at Figure \ref{fig:schedule-study3}.

\subsection{CLIP/LPIPS comparison for stats computation}
\label{app:clip-lpips-comparison}

\begin{figure}[h]
\begin{center}
\subfigure[]{\includegraphics[width=0.48\linewidth]{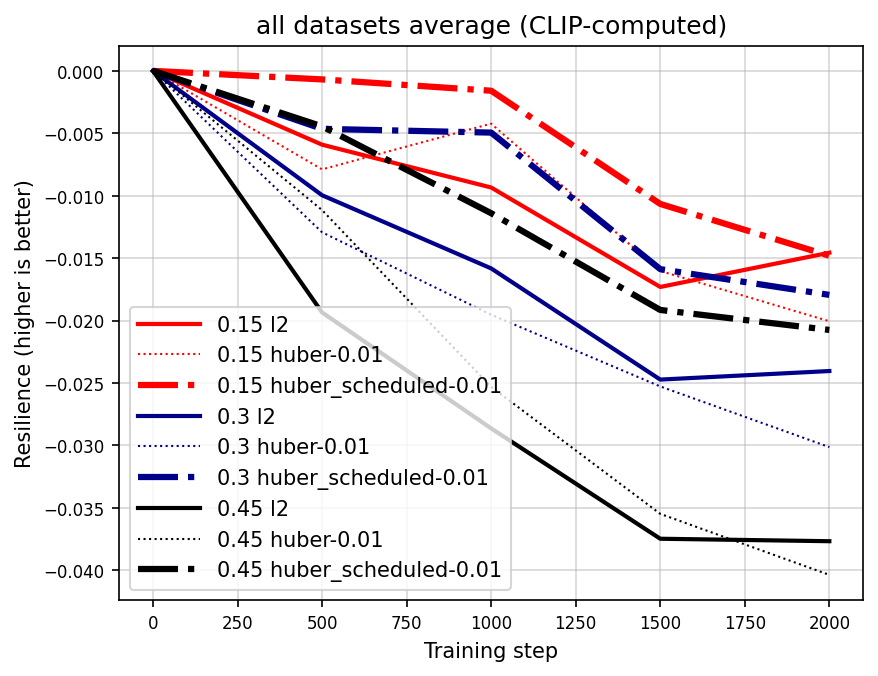}}
\subfigure[]{\includegraphics[width=0.48\linewidth]{images/contamination-dependencies/contamination-depencency-all-datasets-lpips.png}}
\caption{a) Averaged across all datasets CLIP-computed $R$-value at different levels of corruption "0.15 huber\_scheduled-0.1" corresponds to the model trained on the dataset with $15\%$ corruption with Huber loss with exponential decrease to the final value $0.1$ b) The same plot, but computed with \textquote{1 - LPIPS}. LPIPS shows more stability. The fact that the R-value goes above zero in the CLIP 15\% case may be explained by the variance of low number of contamination samples.}
\label{fig:contamination-study1-addon}
\end{center}
\end{figure}

In addition to the \textquote{1 - LPIPS} perceptual metric, we employ CLIP. Figure \ref{fig:contamination-study1-addon} demonstrates that \textquote{1 - LPIPS} score measurements are largely consistent with CLIP. Still, we have ultimately chosen \textquote{1 - LPIPS} as the main evaluation metric due to the reasons stated in the main body of the article.

\subsection{Dependence on the loss constant}
\label{sec:loss-constant}

\begin{figure}[H]
\begin{center}
\subfigure[]{\includegraphics[width=0.48\linewidth]{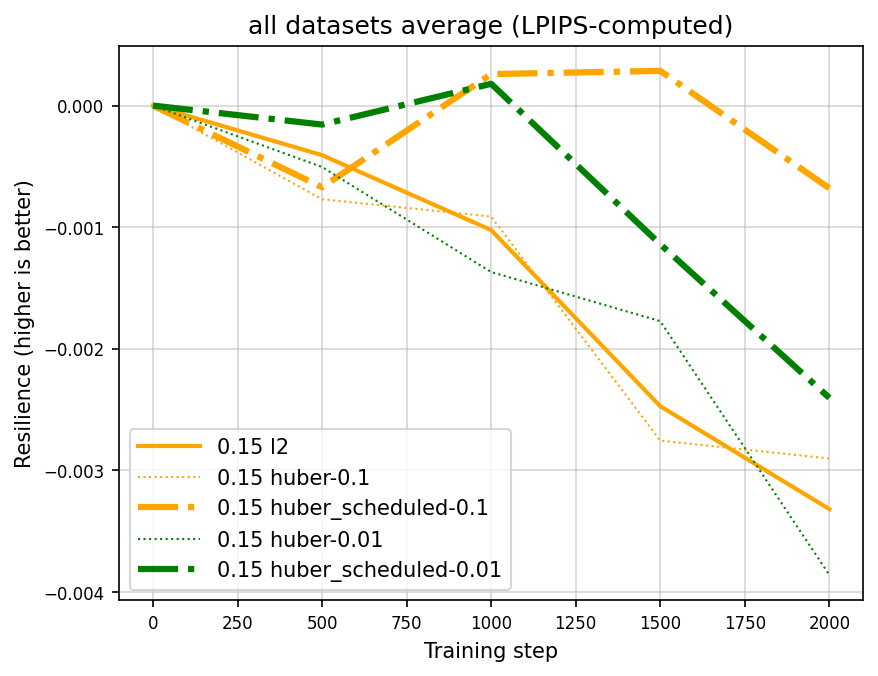}}
\subfigure[]{\includegraphics[width=0.48\linewidth]{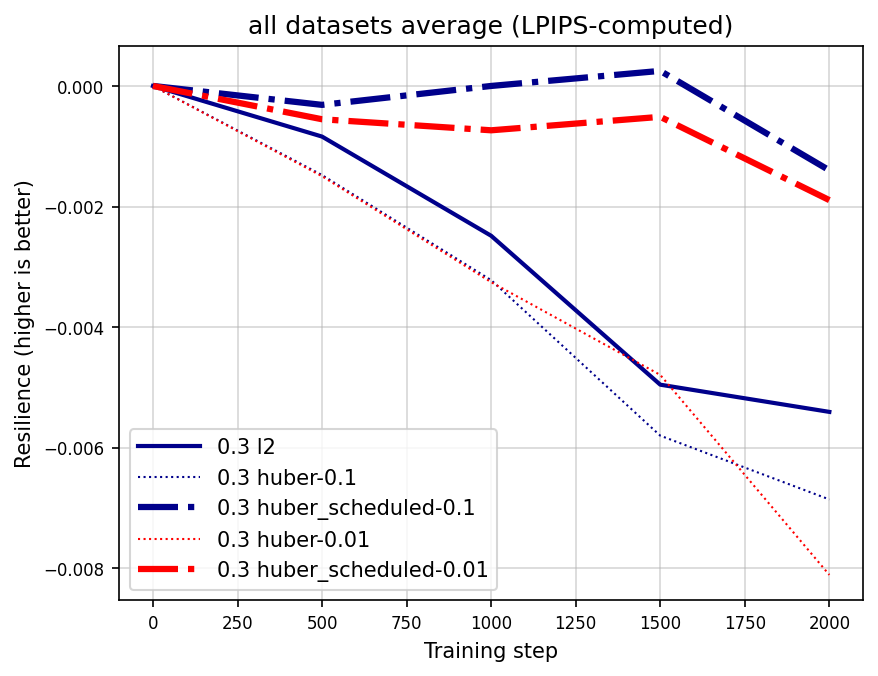}}
\caption{\textquote{1 - LPIPS} $R$-factor for $\delta_0 = 0.01$ and $\delta_0 = 0.1$. and a) 15\% of corruption; b) 30\% of corruption. Higher $\delta_0$ seems to stabilize the results, as the function is closer to L2 on early timesteps.}
\label{fig:c-study-addon}
\end{center}
\end{figure}

For each dataset in addition to L2 training we made two trainings with different Huber loss parameters. As shown at Figure \ref{fig:c-study-addon} increasing the constant improves the stability of the curves by making them closer to L2, although it lowers the robustness potential.

\subsection{All prompts Resilience comparison plots}
\label{app:prompts-comparison-plots}

\begin{figure}[H]
\begin{center}
\subfigure[]{\includegraphics[width=0.48\linewidth]{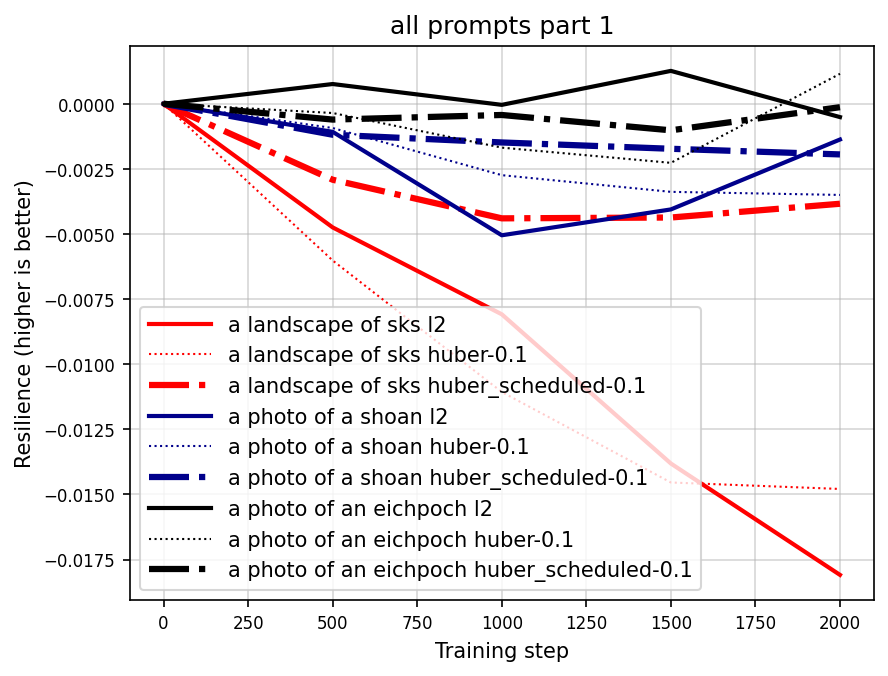}}
\subfigure[]{\includegraphics[width=0.48\linewidth]{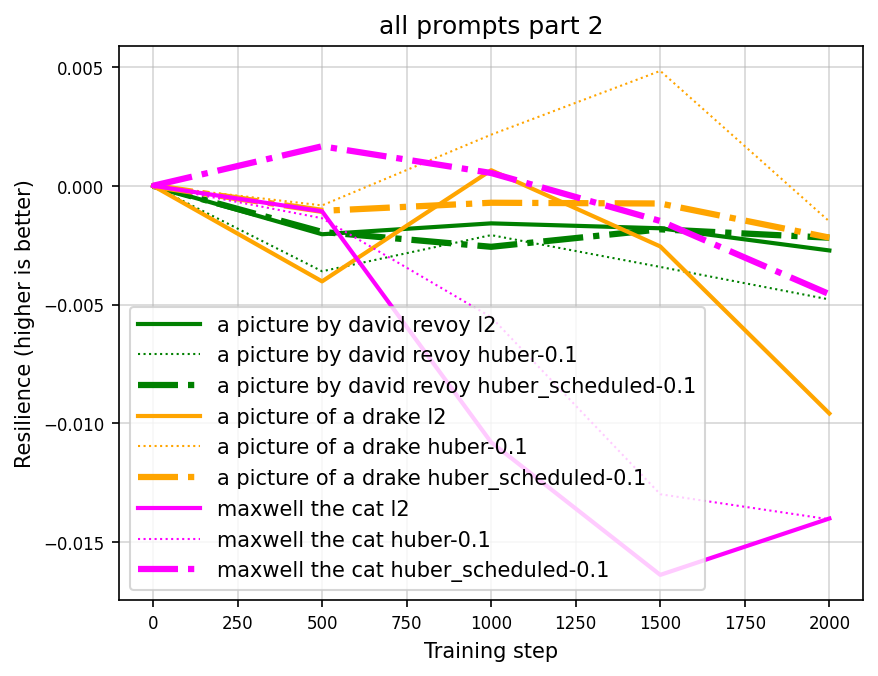}}
\caption{\textquote{1 - LPIPS}-computed Resilience at different training steps on different image datasets/prompts. Corruption level: $45\%$, delta-scheduling: exponential decrease to $0.1$. Final results are reported in the Table \ref{table:prompt-comparison}. Split into two parts for readability. It's visible that these have the very close value ranges, and therefore we have the right to average them for our main-body plots.}
\label{fig:prompts-study-addon}
\end{center}
\end{figure}

Figure \ref{fig:prompts-study-addon} shows the plot of $R$-value at different training steps. The results at the final step are aggregated in Table \ref{table:prompt-comparison}. At this level of corruption, out of 7 datasets our method outperforms L2 in all but one case, though the margin varies.

\subsection{Difference- \textit{versus} Division-based R-factor derivation}
\label{app:r-factor-comparison-plots}

In our experiments, in addition to (\ref{resilience}) we also tried using an alternative formula for the $R$-value, taking into account both the similarity to the clean data and the similarity to the poison:

\begin{equation}
R = S_{\text{to clean}}^{\text{corrupted}} / S_{\text{to poison}}^{\text{corrupted}} - S_{\text{to clean}}^{\text{clean}} / S_{\text{to poison}}^{\text{clean}}
\label{resilience-legacy}
\end{equation}

 While this metric might have been thought to come naturally, where were a number of stability problems resulting basically in inability to derive any statistics comparing different datasets because of the high variance -- making the work look \textquote{cherry-picky} -- and cases causing the factor to unnaturally jump highly above zero. Some of the problems are outlined below:

 \begin{enumerate}
     \item In event of nearly zero similarity to poison the denominator's small value would cause the explosion of the fraction and lead to results greater than zero;
     \item As cosine similarity can be negative, if the similarity to the poison and the clean data both switch sides, the $R$-factor will still be positive;
     \item If the values have proportional positive growth (the similarity to clean data grows and the similarity to the poison grows), the value will stay constant, although it will be more representative if it would come down (examples: $0.5/0.05$, $0.25/0.025$ have the same value).
 \end{enumerate}
 
\begin{figure}[H]
\begin{center}
\subfigure[]{\includegraphics[width=0.48\linewidth]{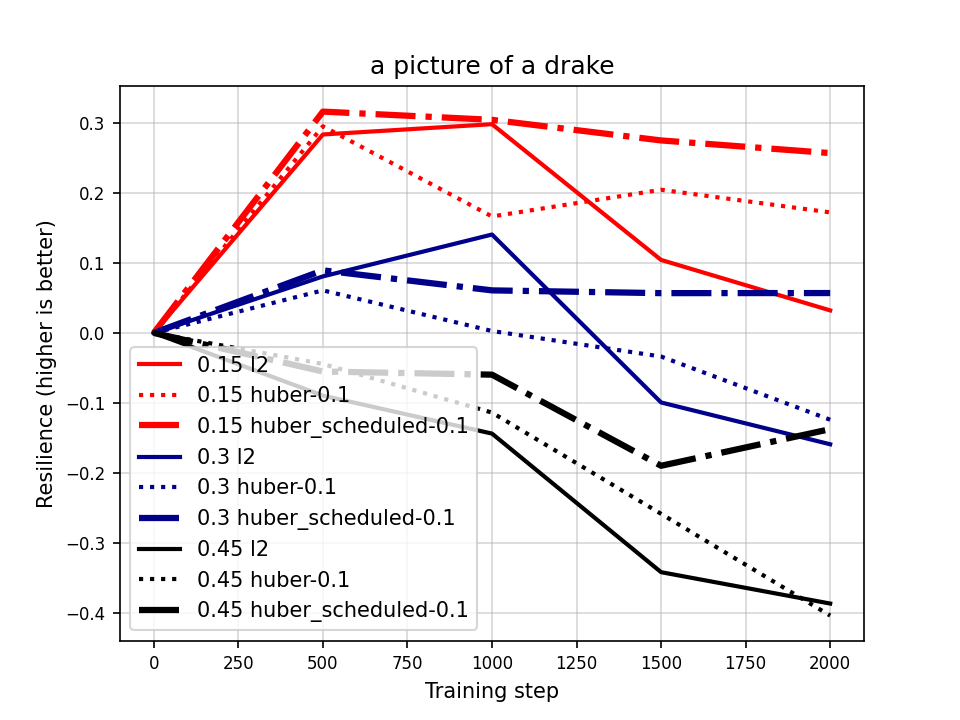}}
\subfigure[]{\includegraphics[width=0.44\linewidth]{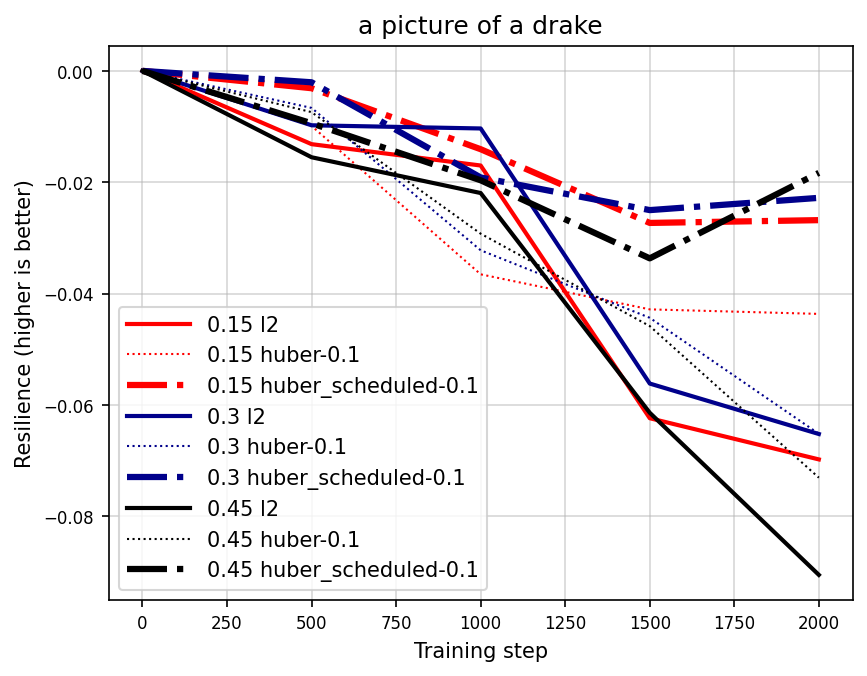}}
\caption{CLIP-computed $R$-value for the prompt "a picture of a drake". The parameters are evident from the plot names: "0.15 huber\_scheduled-0.1" corresponds to the model trained on the dataset with $15\%$ corruption with Huber loss with exponential decrease to the final value $0.1$. The compared variants are: a) Division-based $R$-factor; b) Difference-based $R$-value used in the rest of our work. It's evident that the differential $R$-value's dynamic looks more stable and not above zero. We are showing only one prompt here because of the range instability in the former case.}
\label{fig:r-comparison}
\end{center}
\end{figure}

Therefore it was decided against it. We provide the plots at Fig. \ref{fig:r-comparison} showing how these metrics influence the dynamics and the range of the $R$-values for one example prompt for the CLIP-computed similarity and $45\%$ of pollution.

\newpage

\section{Datasets}
\label{app:datasets}

Below are the samples of the used datasets, one from each. The images are square-resized to 512x512. The number of images in each dataset varies from 6 to 17. All the used datasets, including the automatically composed final versions, on which the trainings have been commenced, are available at the link.\footnote{\url{https://drive.google.com/drive/folders/15JwiRxFYDlFnx5SdT1i78tlZztKNUw3f?usp=drive_link}}

\begin{figure}[H]
\begin{minipage}{\textwidth}
\begin{center}
\includegraphics[width=0.7\linewidth]{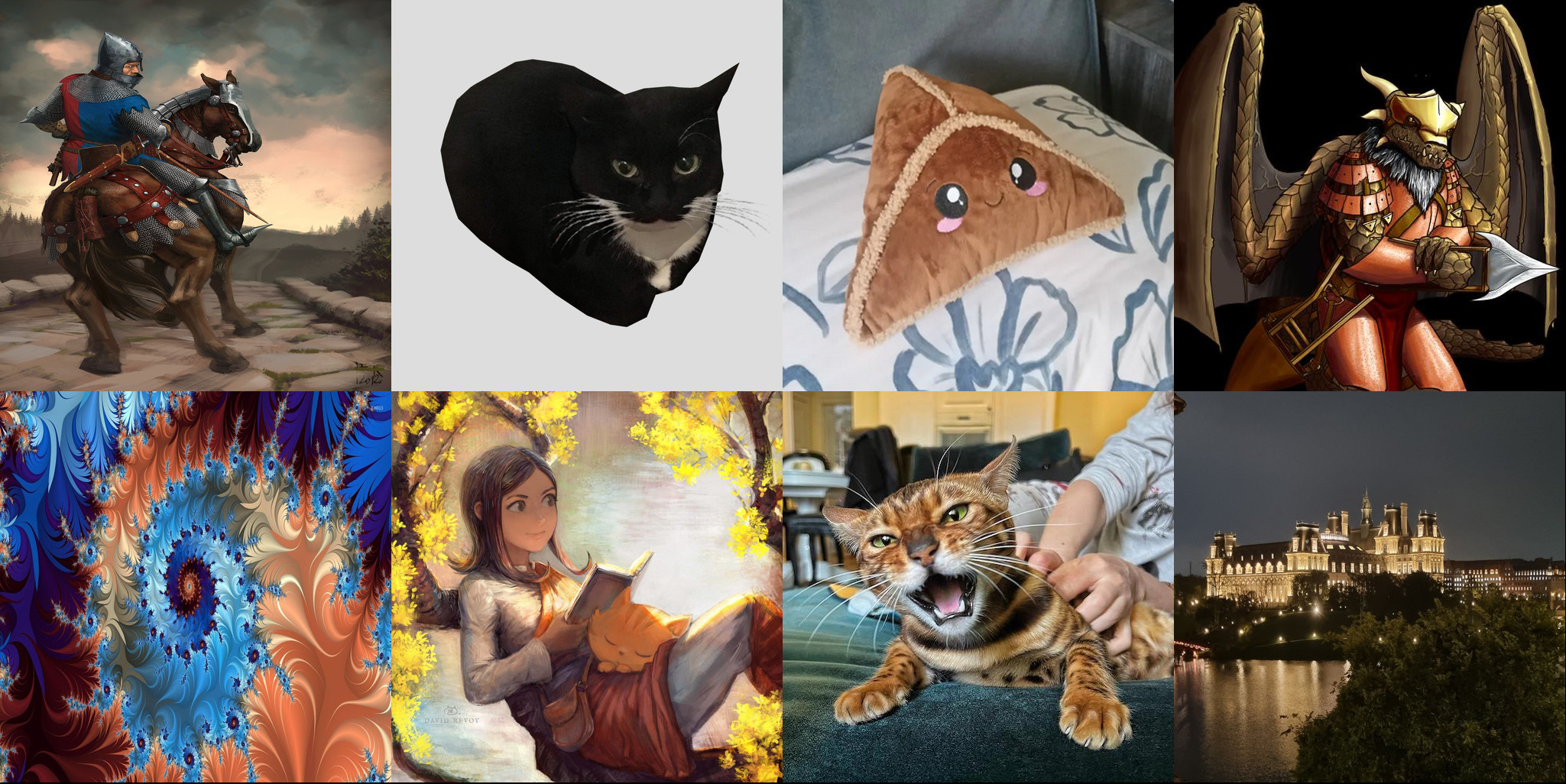}
\caption{From left to right. Upper row: a picture in the style of lordbob\footnotemark, maxwell the cat\footnotemark, a photo of an eichpoch\footnotemark, a picture of a drake\footnotemark. Lower row: random internet picture\footnotemark, a picture by david revoy\footnotemark, a photo of a shoan, a landscape of sks.\footnotemark}
\label{fig:dataset-samples}
\end{center}
\end{minipage}
\end{figure}

\footnotetext{https://www.artstation.com/theartysquid}
\footnotetext{https://knowyourmeme.com/editorials/guides/who-is-maxwell-the-cat-how-this-spinning-cat-gif-became-a-viral-meme}
\footnotetext{https://www.ozon.ru/product/podushka-igrushka-echpochmak-treugolnik-652383324}
\footnotetext{https://github.com/wesnoth/wesnoth/tree/master/data/core/images/portraits/drakes}
\footnotetext{various authors}
\footnotetext{https://www.davidrevoy.com/}
\footnotetext{Contribs listed in \ref{sec:acknowledgements}}

\section{Acknowledgements}
\label{sec:acknowledgements}

The first author wants to say thanks to his friend Ruslan\footnote{https://reddit.com/u/ruhomor} who provided moral support and an extensive dataset of his cat named Shoan photos for training purposes, on which our method ironically fared relatively poorly.

We also thank our friend Ilseyar who made the stunning photos of the castle and the \textit{eichpoch}\footnote{https://en.wikipedia.org/wiki/Uchpuchmak} toy and gave them to be used in our project.

The first author expresses his gratitude to the Deforum AI art community\footnote{https://discord.com/invite/deforum} and its main contributors for being supportive and inspiring.

Additionally, we thank all the artist, photographers and craft makers whose work we used in our experiments.

We greatly thank the reviewers for pointing out the strong and the weak parts of the original version of this paper, helping us to correct it and transition to more objective, numerically, statistically and perceptually stable metrics, while our positive empirical results remained largely intact.

\end{document}